\documentclass[10pt,twocolumn,letterpaper]{article}

\usepackage{cvpr}              
\usepackage{comment}
\usepackage{float}
\definecolor{cvprblue}{rgb}{0.21,0.49,0.74}
\usepackage[pagebackref,breaklinks,colorlinks,allcolors=cvprblue]{hyperref}

\def\paperID{OpenSUN3D-22} 
\def\confName{CVPR}
\def\confYear{2025}

\title{HAECcity: Open-Vocabulary Scene Understanding of City-Scale Point Clouds with Superpoint Graph Clustering}

\author{Alexander Rusnak 
\quad Frédéric Kaplan \\
{École Polytechnique Fédérale de Lausanne}\\
{\tt\small alexander.rusnak@epfl.ch}
}

\begin{document}
\maketitle
\begin{abstract}
   Traditional 3D scene understanding techniques are generally predicated on hand-annotated label sets, but in recent years a new class of open-vocabulary 3D scene understanding techniques has emerged. Despite the success of this paradigm on small scenes, existing approaches cannot scale efficiently to city-scale 3D datasets. In this paper, we present Hierarchical vocab-Agnostic Expert Clustering (\textbf{HAEC}), after the latin word for `these', a superpoint graph clustering based approach which utilizes a novel mixture of experts graph transformer for its backbone. We administer this highly scalable approach to the first application of open-vocabulary scene understanding on the SensatUrban city-scale dataset. We also demonstrate a synthetic labeling pipeline which is derived entirely from the raw point clouds with no hand-annotation. Our technique can help unlock complex operations on dense urban 3D scenes and open a new path forward in the processing of digital twins.
\end{abstract}    
\section{Introduction}
\label{sec:intro}

\begin{figure*}
\begin{center}
  \includegraphics[width=1.0\linewidth]{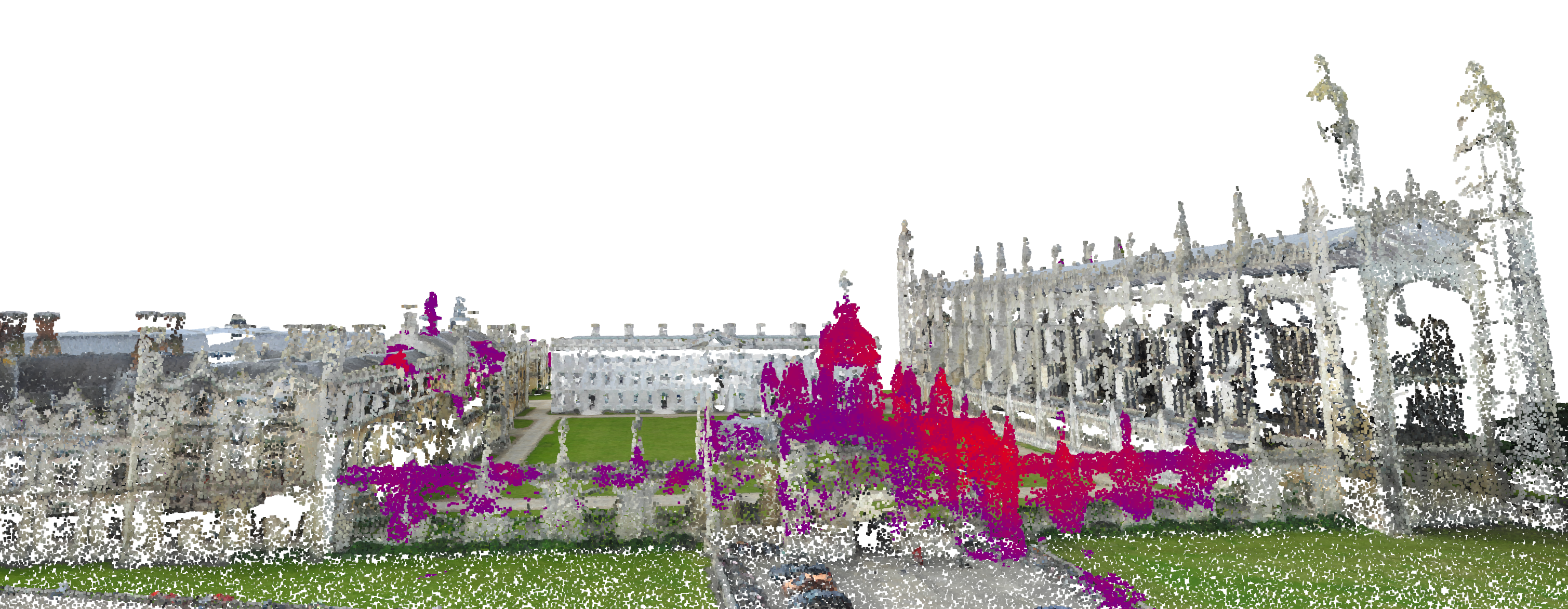}
\end{center}

   \caption{A long-tail query for "The gate of King's College" - we are able to distinguish this singular, particular building despite the architectural similarity of the surrounding buildings and their association with this particular college in Cambridge.}
\label{fig:king_gate}
\end{figure*}

As the barriers for the capture of large-scale 3D models continues to decrease, the importance of scalable, flexible, and effective 3D scene understanding increases in equal measure. To transform large scale 3D data, whether it be digital twins of cities, factories, or aerial scans, from an inert set of millions of points into data infused with semantic meaning is of pressing concern within the field of computer vision. The more traditional approach to 3D semantic segmentation relies on hand-annotated datasets where points are classified into a small number of predefined classes or instances. Though this paradigm has proven powerful in some respects, recent research has approached the problem of open-vocabulary semantic segmentation by utilizing semantic vectors (usually aligned with CLIP \cite{clip_radford2021learningtransferablevisualmodels}) instead of discrete class labels. This paradigm is paramount for large-scale point clouds because  it can unlock self-supervised learning by leveraging foundation model derived pseudo-labels for the 3D scene, thus avoiding cumbersome manual labeling. The relevance of these problems motivates the need for efficient open-vocabulary panoptic segmentation of large scale point clouds. 

Despite the importance of large-scale scenes for many real world applications, the traditional approach to large-scale 3D panoptic segmentation has not received a commensurate level of attention relative to smaller scene formulations of the same problem. This is likely due to the inherent difficulties of this problem: huge scenes contain a substantial number of distinct objects within the millions or billions of points which make them up. Many of the most successful small-scale panoptic segmentation approaches have fundamental limitations which make them unsuitable to processing large scenes. Some rely on high-parameter backbones which can process entire rooms at once but require substantial chunking for large scenes, others have a ceiling for the number of possible object instances in a scene or their approach to instance identification is very memory inefficient and thus doesn't scale. Robert et al \cite{SuperCluster} introduced a highly efficient and accurate panoptic segmentation technique, SuperCluster, predicated on superpoint graph clustering to solve these issues for closed-vocabulary predictions. Though this approach has been successful within its problem setting, there are still fundamental issues with closed-vocabulary models.

The largest downside of closed-vocabulary segmentation methods is the most obvious: the real world is not delineated into a small set of simple object types but in fact contains many unique or long-tail objects which resist easy categorization. Moreover, even within the set of common classes (i.e. chairs) each individual object has many properties that transcend its class label such as color, material, or texture which are crucial for many of the potential use cases of 3D scene understanding. The current state of open-set 3D segmentation research can be primarily delineated across two dimensions. Firstly, by the way in which the multimodal vector labels are derived. Some seek to train language-image-3D embeddings from whole cloth, but most utilize CLIP (or similar foundational models) to derive vectors from the hand-annotated, textual class labels or from the aligned images of the real world upon which the 3D model is based. Secondly, the approaches can be separated into those which create per-point semantic representations, those focused on object-based representations, or those that approach panoptic segmentation of both per-point semantics and instance labels. Panoptic segmentation is especially important because solely instance-level representations can miss important details about objects or subparts of objects, and strictly point-based semantic segmentation can be too noisy with substantial bleed between the labels at the edges of contiguous objects. Another open issue which confounds most semantic segmentation and panoptic segmentation approaches is that it requires substantial memory to store per-point features, especially for large scenes. In order to efficiently compute features for city-scale data, it is thus imperative to use a form of hierarchical representation.

In light of these recognized challenges, we propose HAECcity, which leverages a superpoint graph clustering based mixture of experts (MoE) model trained on data from a novel panoptic pseudo-labeling pipeline to perform efficient open-vocabulary 3D scene understanding on large-scale scenes. We derive the name for this approach as a portmanteau of our model name and city, but also from the medieval scholastic concept of haecceity \cite{dunsScotus1639, sep-medieval-haecceity} which describes the irreducible and specific set of properties that differentiate things into distinct instances of \textit{this particular thing}, i.e. a certain ``thisness" for each ``this." In other words, our model seeks to differentiate not just the specific instances of objects in each class, but the particular characteristics of each instance which differentiates it from the others: not just perceiving that there are chairs in a scene or separating them into countable instances of chairs but distinguishing that this one particular chair is green and made of leather while another is a black desk chair. 

Our contributions can be summarized as such: 

\begin{itemize}
    \item A novel open-vocabulary pseudo-labeling pipeline for point clouds which creates panoptic labels without relying on hand-annotation in any way by utilizing multiple vision foundation models. 
    \item The application of a superpoint graph clustering backbone with bespoke mixture of experts (MoE) blocks to open-vocabulary panoptic segmentation, enabling highly computationally efficient training and inference. 
    \item The first employment of a strictly 3D processing model based, rather than 2D$\rightarrow$3D projection based, open-vocabulary panoptic segmentation system onto city-scale point clouds.
\end{itemize}

In our work, we demonstrate the effectiveness of our approach by comparing its performance in generating instance pseudo-labels on the Scannet (V2) \cite{dai2017scannet} dataset and performing semantic segmentation on the SensatUrban \cite{sensaturban} dataset, without needing to train with any human labels. We aim to step towards a framework which allows for training panoptic segmentation models based on a variety of aggregated datasets which have a diversity of labels or, in many cases, no labels. 

\section{Related Work}
\label{sec:related_work}

\begin{table}[t]

  \small

  \begin{tabular}{l c c c c c}

    \toprule

     \textbf{Approach}& Open-Set & PS & Non-Recon. & Scale & SSL \\

    \midrule

    SuperCluster &  & \checkmark   & \checkmark   & \checkmark  &  \\
    OpenScene & \checkmark  &    & \checkmark   &  &  \checkmark \\
    LERF & \checkmark  &  &  &  & \checkmark \\
    PVLFF & \checkmark  & \checkmark   &  &  & \checkmark \\
    OpenGraph & \checkmark  & \checkmark  &  &  & \checkmark \\
    OA3D & \checkmark  & \checkmark   & \checkmark  & & \checkmark \\
    PanopticRecon & $\sim$ & \checkmark  &   &  &  \checkmark\\
    Search3D & \checkmark  & \checkmark  &   & \checkmark  & \\
    
    Ours & \checkmark  & \checkmark   & \checkmark   & \checkmark  & \checkmark \\

    \bottomrule

  \end{tabular}
  \caption{Comparison of various related approaches for 3D scene understanding. We present that only model which meets the five criteria: open-vocabulary understanding, panoptic segmentation, being a model based approach which processes the 3D directly/singularly, the possibility to scale to city sized datasets, and the possibility to be trained in without any hand-annotated labels.}
  \label{tab:related_work}

\end{table}

There are a number of works which attempt to solve related problems and thus are important to reference. The most relevant antecedents of our research are OpenScene \cite{openscene} and SuperCluster \cite{SuperCluster}. OpenScene is a touchstone in the open-vocab 3D understanding literature: they project features derived from the real images used to capture the scene onto the 3D points within each individual scene before training a `3D distillation' network to predict those features. In order to make classifications, they use an ensemble of the 2D labels and 3D distillation predicted features as the per-point features before selecting a class assignment using cosine similarity between the text embeddings of the class labels and the predicted features. Openscene is able to perform openset queries on Scannet data but has two main drawbacks: it requires calculating 2D$\rightarrow$3D projections for every new scene for its highest performance and it processes every point at full resolution for each scene. Both of these issues prevent the model from being efficient at high scale. On the other hand, SuperCluster is designed specifically for huge scenes and is very computationally efficient in terms of both memory and operations relative to its efficacy. It achieves this performance by clustering points into hierarchical `superpoints' and processing the scene as represented by a graph of these superpoints. Unfortunately, SuperCluster is a closed-vocabulary approach which leverages hand-labeled segmentation datasets and thus is not applicable to our problem of self-supervised training for open-set understanding. The simplest way to understand the ancestry of our modeling approach is as a combination of a modified version of the data preprocessing step of OpenScene and an improved version of SuperCluster for predicting open-set panoptic labels at scale. 

The other research which is comparable to this paper all fall into three general buckets: those which require hand-annotated labels, those which cannot process large scenes due to architectural constraints, and those which are projection or reconstruction approaches requiring heavy computation on every new scene rather than training a model which can classify points in a single pass. Xiao et al. \cite{openvoc-panopt-distill} derive their per-point CLIP features by converting the class label text while Takmaz et al. \cite{search3D} presents an impressive graph-based representation of 2D$\rightarrow$3D projections of CLIP features but unfortunately uses a model trained on ScanNet labeled data in order to generate their instance segmentations. UniSeg3D \cite{unified} demonstrates SOTA performance across multiple scene understanding tasks and is a promising development for performing various tasks with single models, however, because they process the scene as single-level superpoints rather than using a hierarchical or reduced dimensional representation of the scene, their approach does not scale to large scenes. LERF \cite{lerf}, PVLFF \cite{pvlff}, OpenAnnotate3D \cite{zhou2023openannotate3dopenvocabularyautolabelingmultimodal}, OpenGraph \cite{opengraph}, Li et al. \cite{li-open-dense}, and PanopticRecon/ PanopticRecon++ \cite{yu2024panopticreconleverageopenvocabularyinstance, panoptic-recon++} all allow for some or all of the desired functionality on 3D scenes but each is a reconstruction or projection approach which requires each new scene to be expensively preprocessed. For PanopticRecon, there are no examples of someone using the backbone of their approach, instant neural graphic primitives \cite{iNGP}, to reconstruct cityscale scenes, making its feasibility doubtful for our task. We also considered basing our training data feature projection on PanopticRecon, however their approach is only ostensibly open-vocabulary: they require a set of ``interested classes" text labels which are a priori relevant before using Grounded SAM \cite{ren2024groundedsamassemblingopenworld} to produce masks for just those classes. In order to potentially extend this labeling backbone to a true open-vocabulary approach, we tested taking the vocab-agnostic predicted CLIP features produced by Semantic SAM \cite{li2023semanticsamsegmentrecognizegranularity} (a closely related but faster approach to Grounded SAM which uses a unified model) but found them to be less effective than the mask-level vision-language features produced by OpenSeg \cite{openseg}. 

We present the comparative benefits and limitations of the various similar approaches in Table \ref{tab:related_work}.

\section{Method}
\label{sec:method}

Given a 3D point cloud \(P\), our goal is to train a model which predicts CLIP features $\mathbf{f}^C$ and an object index label $\text{obj}(p) \in \mathbb{N}$ for each point \(p \in P\). In this section, we first detail our pseudo-labeling pipeline which produces semantic CLIP features through 2D$\rightarrow$3D projection before clustering on those 3D CLIP features to produce instance labels. Afterwards, we describe our backbone model which modifies the SuperCluster approach to apply to open-set rather than closed-set panoptic segmentation. 

\subsection{Semantic Projection and Pipelining}
In order to create a set of CLIP features for a 3D scene \(P\), we need a set of pose-aligned images which depict the scene. For many datasets, such as SensatUrban, there are no available source images and it is necessary to generate a set of synthetic images derived from casting cameras into the 3D scene. Following this necessity, we capture synthetic images of our scenes by initializing a 3D grid of points equally spaced within the scene which all fall within a margin from the edges of the scene. We then generate camera extrinsic matrices as if we placed eight cameras at each point, arranging them in the four cardinal directions as well as diagonally. By passing these extrinsics to an Open3D renderer \cite{open3d}, we create a set of images of the 3D model from each view. This leaves us with a set of synthetic RGB images for each scene, \(I\),  with a resolution of $H \times W$.  

When casting synthetic cameras into the scene, it is possible that one of them falls inside of an object in the point cloud or another strange location, and thus would have a nonsensical, occluded view. In order to rectify this, we validate image quality using MobileCLIP \cite{mobile_clip_Vasu_2024_CVPR}. We take a set of positive labels (i.e. ``a normal scene", ``an indoor scene", ``an outdoor scene") and negative labels (i.e ``an incoherent image", ``unorganized, random points", ``a blank image") which we convert into text embeddings. We find the cosine distance between the embeddings for each image and the class labels before discarding any image which is not assigned to a positive class with a greater than 65 percent probability. 

After performing this sanity check, we determine if a sufficient number of points in the scene have a corresponding image. We can easily project the points onto the 3D scene by calculating the corresponding pixel $u$ for each point $p \in P$. Since we have the camera intrinsic $I_i$ and extrinsic matrix $E_i$, we can define $\tilde{u} = I_i \cdot E_i \cdot \tilde{p}$. To ensure that we are mapping the pixels only to visible points which feature prominently in the scene, we also perform a depth screening. We use the Depth Anything Model \cite{depthanything} on each of the approved images to produce synthetic depth maps and discard any corresponding pixel which is beyond a depth distance threshold. 

After the initial mapping of pixels back to the scene, there is often a substantial number of points without a paired pixel. To rectify this, we calculate clusters on the unmapped points using adaptive DBSCAN \cite{dbscan, adbscan} and select two points randomly from each cluster. This results in a small subset of points that, when photographed, have a high coverage of the unlabeled points because many of them are either directly adjacent or in the same areas of the scene. We cast eight cameras into the scene looking at each point in the subset from various angles (the corners of a cube), and perform the same sanity check on these images before mapping them back to the scene. We repeat this step recursively until it achieves a high image coverage of the scene.

The next section of our feature projection approach contains two primary steps: image feature extraction and 2D$\rightarrow$3D projection. We use OpenSeg \cite{openseg} to compute per-pixel embeddings for each $i \in I$ which leaves us with the feature set $\mathbf{I}_j \in \mathbb{R}^{H \times W \times C}$ in which $j$ is the number of images and $C$ is the feature dimension. With this feature set, we project the features into the scene as described above. In many cases, there are multiple corresponding pixels for a single point. In these cases, we simply average the assigned CLIP vectors. After passing the 3D scene through this pipeline we are left with a relatively dense set of points with assigned feature vectors corresponding to 
\begin{align}
\mathbf{f}(p) = \frac{1}{|\mathcal{J}_p|} \sum_{(j,i) \in \mathcal{J}_p} \mathbf{I}_j(i)
\end{align}

In order to create instance labels without manual annotation, it is necessary to group the points into sets of objects using the semantic vectors we have projected into the scene. To achieve this, we first utilize spherical k-means, which has been shown to be an effective method for semantic clustering of CLIP vectors, to assign the points into groupings with a similar conceptual valence \cite{ spherical3, spherical1, spherical2}. We consider these groupings pseudo-class labels for each given scene, and take an average of the constituent CLIP vectors as the representative vector for that pseudo-class. Given these semantic point groupings, we further subdivide them within each class by clustering over the point positions using adaptive DBSCAN where the epsilon and minimum points hyperparameters are set relative to the density of the points in each class set. We consider these groupings as object instances: 

\vspace*{10pt}

\textbf{Step 1: Pseudo-class labels $z_{\text{pc}}$ from spherical k-means clustering:}
\begin{align}
z_{\text{pc}}(p) = \underset{k \in \{1,\ldots, K\}}{\operatorname{arg\,max}} \; \mathbf{f}(p)^\top \mathbf{u}_k,
\end{align}

\textbf{Step 2: Pseudo-instance Labels $z_{\text{pi}}$ from DBSCAN within pseudo-classes:}

\begin{align}
\mathcal{P}_k = \{\, p \in \mathcal{P} : z_{\text{pc}}(p) = k \,\}, \quad k = 1,\dots,K
\end{align}

\begin{align}
{z_{\text{pi}}(p) = \operatorname{DBSCAN}\!\Bigl( \{ \mathbf{x}(q) : q \in \mathcal{P}_{z_{\text{pc}}(p)} \};} \nonumber \\ \varepsilon_{z_{\text{pc}}(p)}, \, \text{minPts}_{z_{\text{pc}}(p)} \Bigr )
\end{align}
\vspace*{3pt}

\begin{figure}[t]
\begin{center}

   \includegraphics[width=0.8\linewidth]{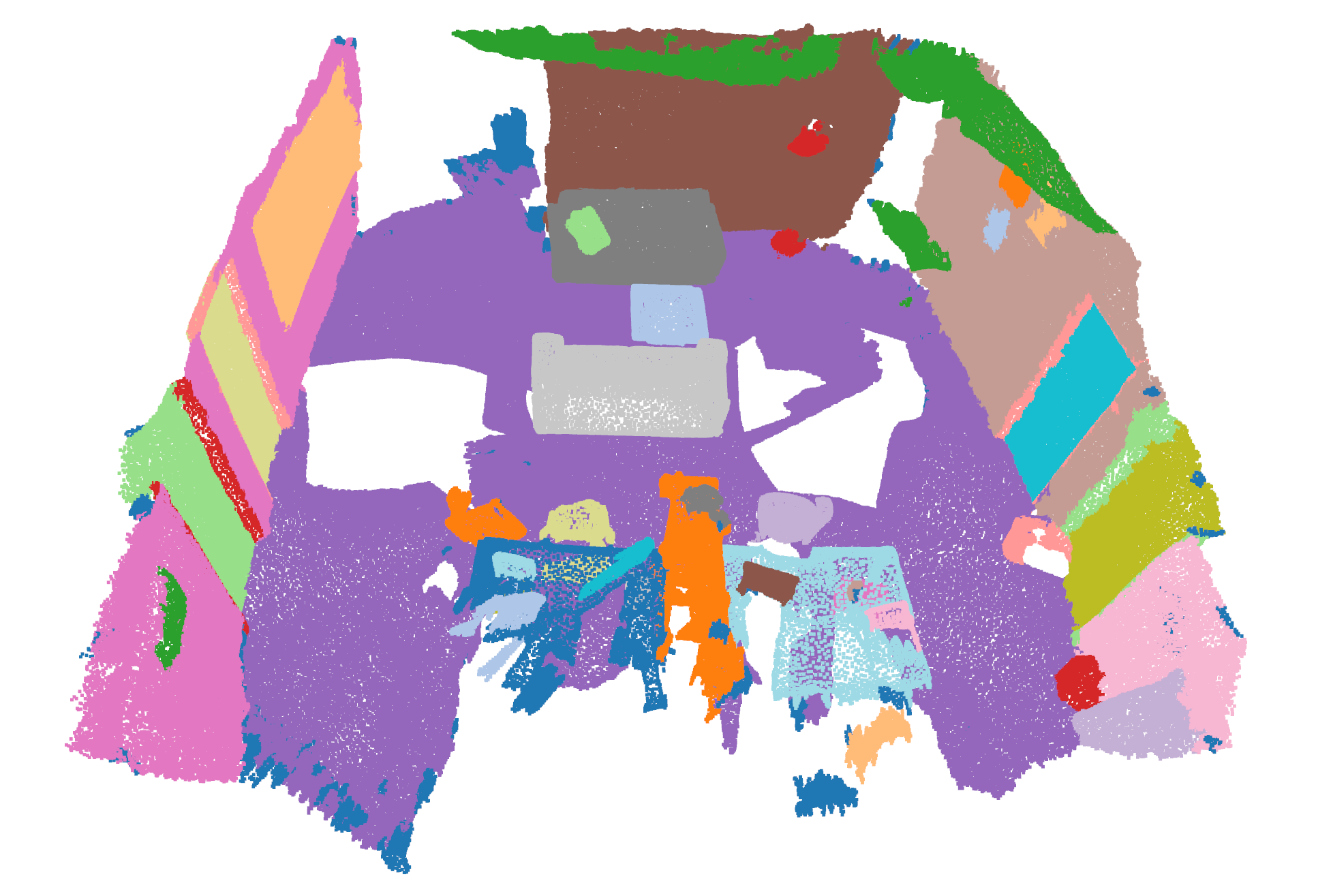}
\end{center}
   \caption{Hand-annotated instance labels on a scene from the Scannet dataset.}
\label{fig:handannotate}
\end{figure}

\begin{figure}[t]
\begin{center}

\includegraphics[width=0.8\linewidth]{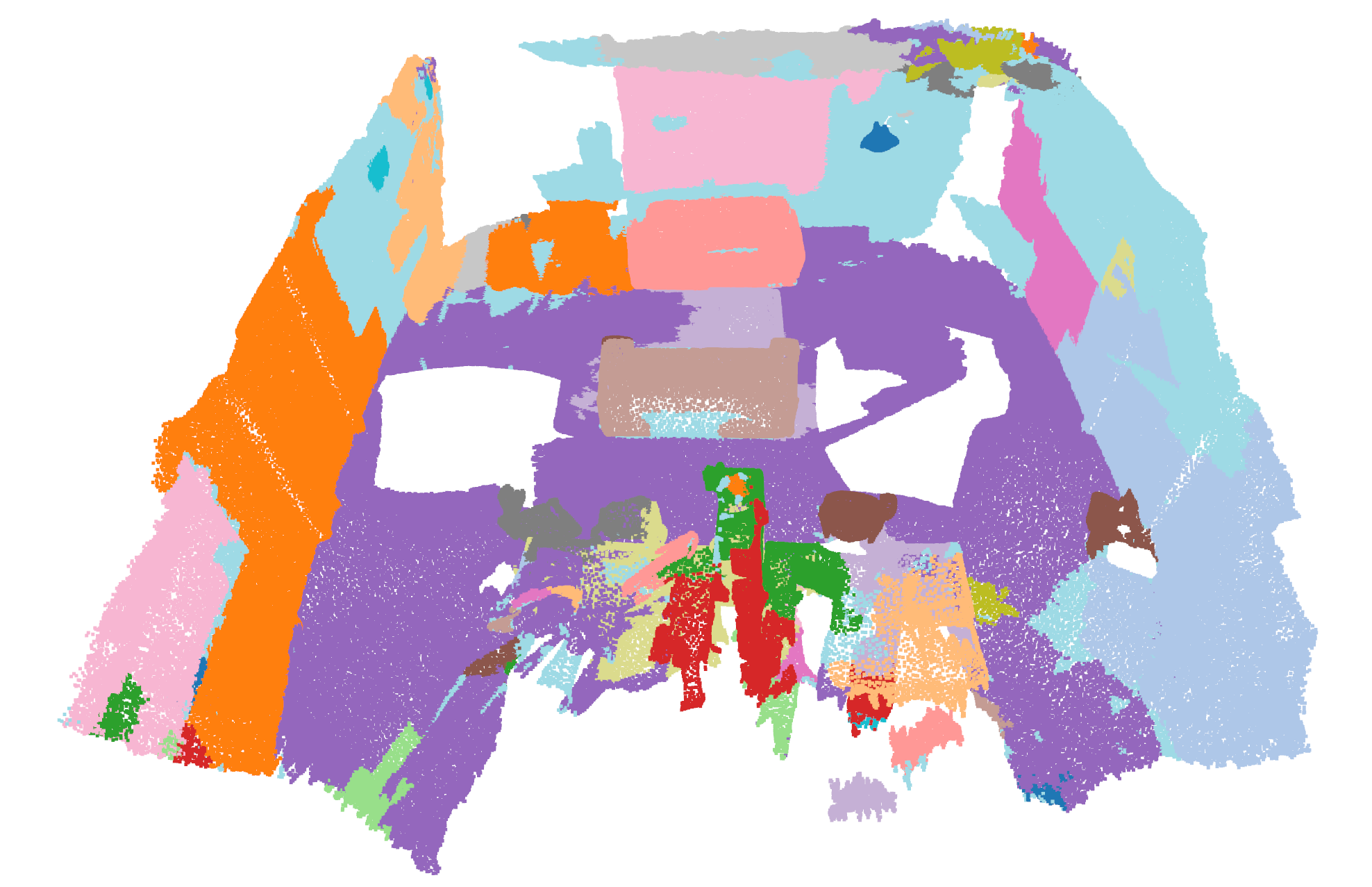}
\end{center}
   \caption{Feature clustering based pseudo instance labels on the same Scannet scene.}
\label{fig:pseudo_instance}
\end{figure}

Though the pseudo-instance labels can be fuzzy or sometimes capture distinct parts of objects rather than what a human would label a whole object, in aggregate they provide a strong basis for self-supervised instance label targets. As demonstrated in Figure \ref{fig:handannotate} and Figure \ref{fig:pseudo_instance}, there is often strong agreement between the hand-annotated and pseudo-labels; it is also apparent that the pseudo-labeling has distinguished between the unique instances of couches and chairs within the scene.

It is common practice in panoptic segmentation tasks to differentiate classes into ``things" (i.e. a chair or car) which are countable, distinct objects for which we would like to predict instance IDs, and ``stuff" (i.e. floor, grass, or water) which are amorphous and unimportant to count. In order to make this distinction, we once again perform an abstract class prediction using cosine similarity with the positive class being ``an object" and the negative class being ``amorphous, uncountable stuff". This allows us to make a distinction between the two types of instances without being overly specific about the things / stuff classes, and thus to avoid allocating computation resources to subdividing sections of a field or the like where only semantic labels are necessary.

\subsection{Model}
\label{sec:model}

Our modeling approach is based on the superpoint transformer (SPT) architecture \cite{superpoint}, particularly within a SuperCluster framework for panoptic segmentation \cite{SuperCluster}. Though it is beyond the scope of this paper to describe every detail of the SPT architecture, it is necessary to have a general overview in order to understand our own innovations. Many point cloud processing models either process every individual point or perform a grid-based sampling of points (voxelization) in order to reduce the number of points to compute over. Superpoint techniques attempt to partition the points into subsets which are responsive to the complexity of the geometric structure such that they are representative of the layout of the scene. Ideally the points making up a simple but large planar structure like a wall would be represented by the same superpoint, whereas a complex object such as a desk would be grouped into multiple distinct superpoints. During preprocessing, SPT partitions the raw points into superpoints by solving an energy minimization problem on a spatial adjacency graph of points where the features are the point's local geometric and radiometric information. This approach is applied recursively on the superpoints, creating a hierarchical graph representation of the point cloud where the deeper levels are strictly more coarse (and thus contain fewer points).

With these hierarchical superpoints, SPT uses a graph-oriented sparse self-attention mechanism which considers the representation produced by the prior level and the edge attributes between superpoints at each scale to learn local scene representations. The processing of the partitions proceeds from fine to course partitions with an encoding step after each partition level, and then inversely from coarse to fine when decoding. This reduces the complexity of performing pointcloud operations across large-scale scenes substantially in terms of memory usage of the scene representations - it also means that SPT has a smaller parameter count than other similarly performant point cloud processing architectures in addition to training significantly faster. 

SuperCluster is a derivative of the SPT approach which frames panoptic segmentation as a graph clustering problem on the superpoints. It utilizes a single SPT backbone to create superpoint level scene representations which are passed to two heads: one for semantic classification and one for predicting the object affinity between superpoints which are represented as edge weights in the point adjacency graph. By adjoining the superpoints which have sufficiently high affinity predictions it is possible to produce instance labels, and thus panoptic segmentations when combined with semantic predictions from the other head. Though this approach shows strong performance across multiple datasets at various scales, its relative superiority to competitor models is most apparent on large scenes. 

Our primary innovations to the original SuperCluster framework were in converting the model to a MoE where each expert at each level is the graph attention module from the original SPT approach, as well as modifiying the semantic head and loss functions. Additionally we rebuilt the superpoint partitioning scheme for the target variables in order to handle CLIP vectors in conjunction with the traditional labels when available. 

The backbone of our Hierarchical vocab-Agnostic Expert Clustering approach is the same as the original SPT in regards to the initial encoding block, relative positional encoding, and general hierarchical encoder-decoder structure. However, we modified the foundational transformer blocks which are used to down and upscale the representation graph between each superpoint partition level in order to increase representational flexibility / parameter capacity while maintaining speed by introducing top-k expert gating based on the representation of each superpoint and the mean-aggregated relative positional encodings for incoming edges to each superpoint node \cite{NEURIPS2022_moe, wang2023gmoe}. Given the per-node routing output, we compute a load balancing loss on the gating probabilities before selecting the top-2 experts to pass the node and graph data through.

Instead of outputting a class prediction from the semantic head, we instead use the semantic head to predict a CLIP vector for each superpoint - furthermore we modified the architecture of the semantic head by adding multiple linear layers in addition to GELU \cite{gelu} activations in order to more gradually increase the size of the representation vector derived from the HAEC backbone. 

When propagating the CLIP features down the partitioning hierarchy, we take a multifaceted approach. Within each superpoint partition at the finest level, we use spherical k-means to cluster across the features and select the cluster centers from the two most populous clusters as the representative values of that superpoint. Simultaneously, we take the representative vector from the most frequent pseudo-class in the superpoint, leaving us three separate vectors for each superpoint at the first level. For all subsequent levels, the selection criteria for the third representative vector is the same. For the two cluster derived vectors, the same spherical k-means clustering is performed on each vector individually and the single most populous vector is selected. So, when applying the loss functions, the similarity is calculated from the prediction to all three target vectors and the average is taken as the loss value at that particular superpoint. 

We utilized two separate loss functions relative to the target CLIP vectors: a reconstruction loss which is simply cosine similarity between the target vector $v$ and predicted vector $\hat{v}$:

\vspace*{10pt}
\textbf{Cosine similarity reconstruction loss:}
\begin{align}
L_{\text{rec}} = 1 - \frac{\hat{v} \cdot v}{\|\hat{v}\| \|v\|}
\end{align}

As well as a triplet loss which leverages our semantic clustering based CLIP pseudo-class labels:

\vspace*{10pt}
\textbf{Triplet Loss (using cosine similarity with margin \(\alpha\))}:

\begin{align}
L_{\text{triplet}} = \max \Big\{ 0, \; \cos\big(\hat{v}_a, \hat{v}_n\big) - \cos\big(\hat{v}_a, \hat{v}_p\big) + \alpha \Big\}
\end{align}

Regarding the triplet loss, the possible positive samples for any superpoint are those with the same pseudo-class, whereas the negative samples are all those from the other classes. During training, we randomly sample one positive $\hat{v}_p$ and negative sample $\hat{v}_n$ relative to each anchor at each iteration. For each superpoint, we minimize the cosine similarity between the predicted representations of the anchor and positive sample while maximizing it for the negative sample. In practice, this creates a productive tension between the granular specificity of the reconstruction loss and object level focus of the triplet loss which helps the HAEC model produce better feature predictions.

\section{Experiments}
\label{sec:experiments}

\begin{table*}[t]

    \centering

    \caption{Panoptic Segmentation Quality Using Different Methods on Indoor and Outdoor Datasets}

    \begin{tabular}{l|ccc|cc}

    \toprule

    & \multicolumn{3}{c|}{\textbf{ScanNet(V2)}} & \multicolumn{2}{c}{\textbf{SensatUrban}} \\
    \hline

     & PQ & mIoU & mAcc & mIoU & mAcc \\

    \midrule

    \multicolumn{3}{l}{\textbf{Closed-vocabulary:}} \\
    
    SuperCluster

        & 58.7 & 66.1 & 

        &  &   \\

    RandLA-Net

        &  & 64.5 & 

        & 69.64 & 52.69  \\

    \midrule

    \multicolumn{3}{l}{\textbf{Open-vocabulary:}} \\

    OpenScene

        &   & 47.5  & 70.7 

        &  &   \\

    Preprocessing Oracle

     &  &  &  

     & 5.62 & 17.65 \\

    HAEC (Ours)

        & 40.03  & 50.62 & 71.93 

        & 22.45 & 30.38  \\

    \bottomrule

    \end{tabular}%

    \label{tab:panoptic_quality}

\end{table*}

In order to validate the 3D scene understanding ability of our model, we evaluate performance on a traditionally annotated closed-set semantic segmentation dataset. After comparing our performance on traditional benchmarks, we demonstrate the ability of our model to query long-tail or unique objects in city-scale scenes.

\subsection{Datasets}
\begin{figure}[t]
\begin{center}
   \includegraphics[width=0.99\linewidth]{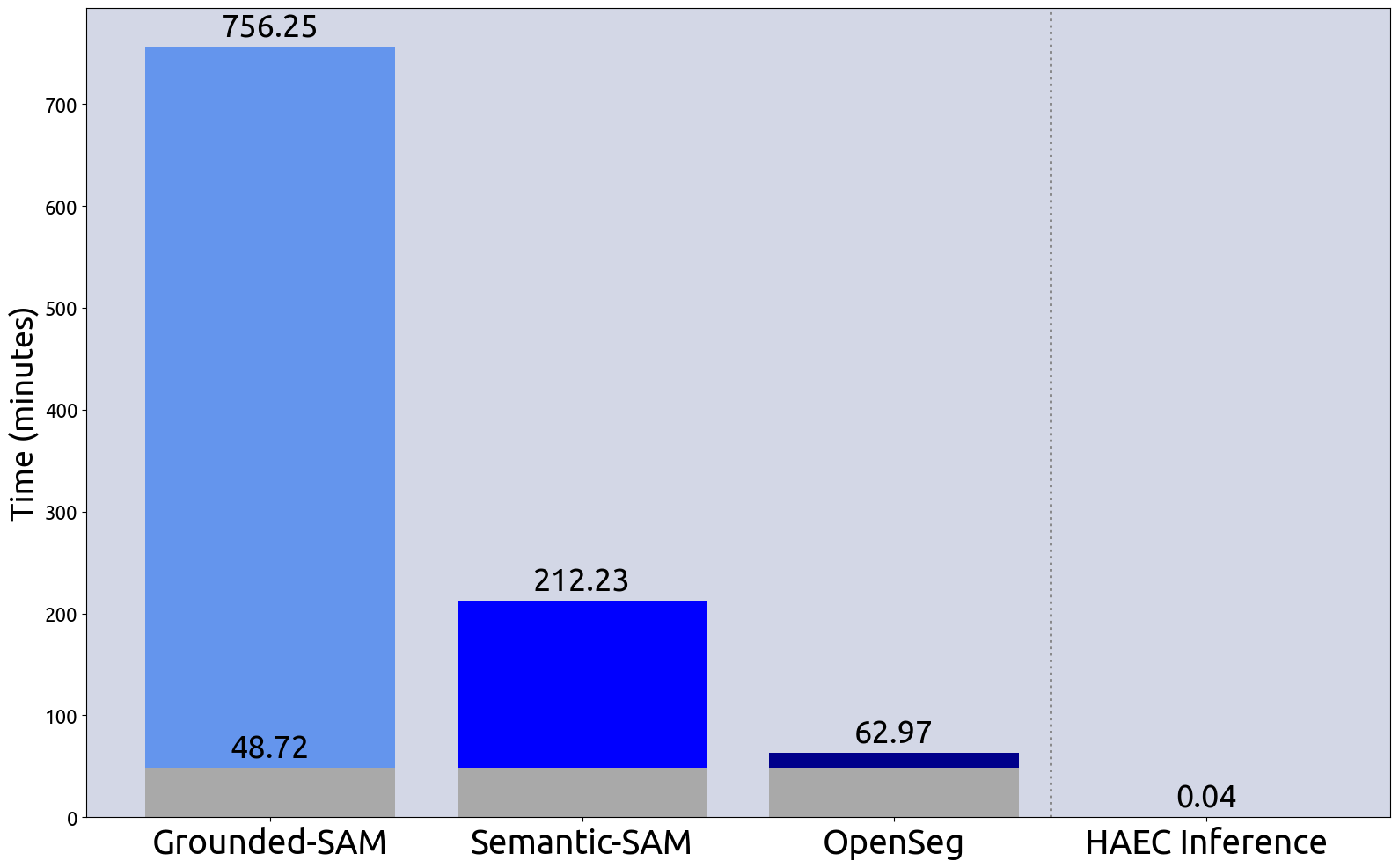}
\end{center}
   \caption{Comparison of estimated processing speed for a scene with 15k precomputed images by the the image annotation method used. The grey section of the bars represents the time for the image to point projection mapping computations. HAEC inference occurs solely on the 3D data of the same scene, whereas these the slow preprocessing steps must be computed for every novel scene when using a reconstruction based approach. The bar for OpenSeg is representative of the preprocessing time for training samples in our approach once the synthetic images are captured. }
\label{fig:speed}

\end{figure}
We used multiple datasets to evaluate our approach which are listed here:

\begin{itemize}
    \item ScanNet \cite{dai2017scannet}. This dataset consists of 1501 medium-scale indoor scenes and is a fairly common benchmark for 3D scene understanding. We evaluate our instance pseudo-label quality and model performance on this dataset on a validation set of 10 percent of the scenes. 
    \item SensatUrban \cite{sensaturban}. This dataset present an urban-scale photogrammetric point cloud dataset with nearly three billion annotated points. This dataset consists of large areas from two UK cities, covering about 6 km² of the city landscape. We report results on a subsampled validation set of 10 percent of the scenes in the dataset. Unfortunately, this dataset does not include instance labels and is thus only evaluated for semantic prediction. 
\end{itemize}

\subsection{Panoptic Segmentation Performance}

Though there is no true peer model for our approach because of the limitations listed in the related work section, it is still useful to compare the relative performance. Despite our approach trailing in some metrics to either closed-set prediction models or reconstruction oriented pipelines, it is by far faster than reconstruction oriented approaches. The PanopticRecon authors acknowledge this drawback by claiming their approach ``does not prioritize real-time performance." In Figure \ref{fig:speed}, we present the comparative time for scene preprocessing based on various image annotation models and our point to image projection technique in grey. Since the PanopticRecon authors have not published their code, we cannot run their approach directly. However, we can assume that the Grounded-SAM image processing time forms a lower time bound for their approach because they pass each image through that model. The bar utilizing OpenSeg is representative of the preprocessing time for OpenScene and HAEC. Both PanopticRecon and OpenScene utilize the results of these preprocessing steps at inference time, while HAEC is applied to the raw point cloud directly and thus is massively faster. In Table \ref{tab:panoptic_quality} we present the metrics on Scannet for panoptic segmentation and SensatUrban for semantic segmentation. For the Scannet results, we utilized the base 2D-3D projections provided by the Openscene authors, and applied our instance-pseudo labeling pipeline and HAEC model to the scenes in order to specifically validate those subsets of our approach. For SensatUrban, though our metrics trail RandLA-Net \cite{hu2019randla}, our problem formulation is substantially more difficult because it is open-vocabulary and suffers from the quality of the pseudo-labels which can be derived from natural images vs synthetic images. To illustrate this difficulty, we have included the performance of a preprocessing oracle (i.e. how accurate the labels derived solely from our pseudo-labeling pipeline are on our validation set). To our knowledge, we are the first to apply any form of open-vocabulary understanding to the SensatUrban dataset. 

\subsection{Open Querying of Unique Objects}

\begin{figure*}
\begin{center}
  \includegraphics[width=1.0\linewidth]{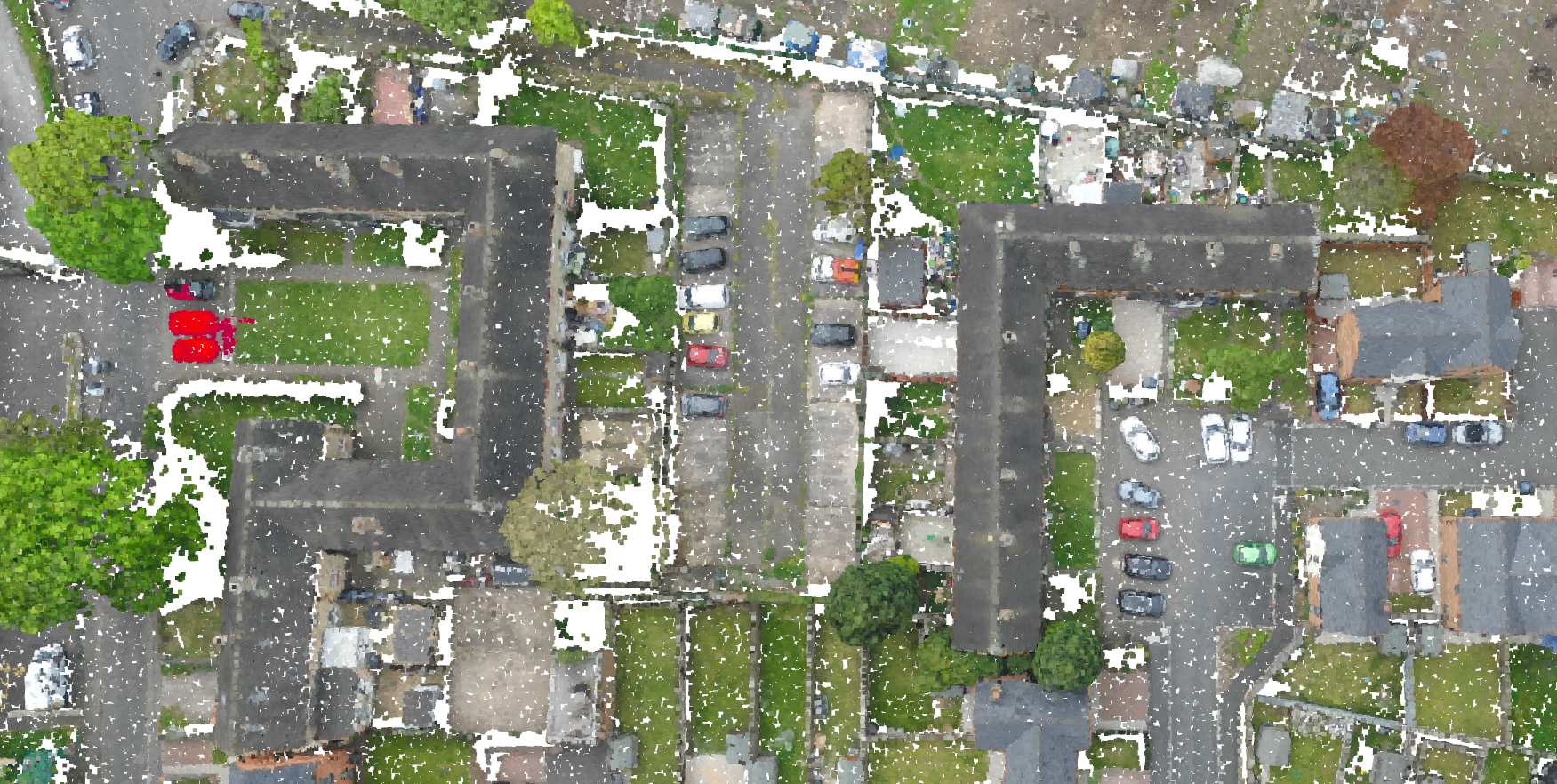}
\end{center}
   \caption{A query for ``a red car" in one of the Birmingham scenes, red indicates a higher similarity and natural colors indicate that a point is below the classification threshold.  }
\label{fig:car_query}
\end{figure*}

 The most important benefit of open-vocabulary 3D understanding models is the ability to recognize long-tail or unique objects which do not appear in traditional semantic segmentation class labelsets. In order to demonstrate this ability and highlight some of the limitations of our approach, we query the preprocessing derived point cloud features on two scenes from SensatUrban. We demonstrate in Figure \ref{fig:king_gate} a query for a famous and particular building: the King's College gatehouse. We are able to find this building despite the bevy of similar buildings nearby, including others (such as the King's College chapel) which are equally famous and likewise associated with Cambridge / King's College. The quality of the features for this particular building are likely attributable to long sight lines surrounding it which allows for higher quality synthetic image coverage (and thus feature coverage) of these points. We elaborate on the issue of feature coverage in Figure \ref{fig:car_query} - it is clear that in some places the features can accurately identify the ``red cars" (on the left of the scene) indicating that the synthetic feature representations are accurate enough. However, in some cases, cars of the correct red color are missed altogether, indicating that there is an uneven coverage of high quality cameras views of the scene. We discuss possible mitigating solutions to these problems in the following section.

\section{Limitations and Future Work}
\label{sec:limitations}

The most significant limitation to our approach comes from the quality of the synthetic-image-derived CLIP features and thus, downstream in the preprocessing pipeline, the per-point CLIP labels. We found our model achieved comparable performance on Scannet to OpenScene when the training data features were derived from real rather than synthetic images taken from the same poses as the real cameras views, pinpointing the vision-language model as an issue. The OpenSeg image encoder is likely unfamiliar with these types of point cloud images as its training dataset is made up strictly of real images. A future direction of research to improve these image-derived feature representations is to finetune a VLM to minimize the embedding distance between the real images of scenes and the corresponding synthetic images in order to improve its performance on this unique type of image. 

Our model is also limited by the superpoint partitions in achieving the highest possible levels of semantic accuracy. Though this process reduces computational complexity, it also makes it harder to learn the most fine-grain information about our source data. Though in the original supercluster paper they demonstrate that the ceiling of this approach is still quite high, the propagating of semantic vectors through the encoder-decoder hierarchy is more complicated than the propagation of simple class labels. 

Another substantial limitation is the high memory and compute consumption of our preprocessing approach which requires us to subsample or subdivide larger scale point clouds into chunks despite the capability of our downstream HAEC model of handling larger point clouds. Though the preprocessed point clouds are still of considerable size, a future goal for this research is to increase the size of the point clouds which can be preprocessed directly with a reasonable time complexity. This could be achieved by changing the camera sampling approach to better capture the scene with fewer redundant or poor quality images. The projection of the feature vectors could also occur at the first superpoint level rather than at the raw point cloud level, which would have a similar effect on complexity as substantially lowering the number of points in the scene. Once these improvements in computational efficiency are achieved, it will be easier to train our approach across multiple large scale datasets concurrently to realize the potential of this approach as a foundational model for city-scale point clouds.

\vspace*{-1pt}
\section{Conclusion}
\label{sec:conclusion}

In this paper we introduced a novel open-vocabulary pseudo-labeling pipeline which creates panoptic labels without relying on hand-annotation. Using those pseudo-labels, we applied our Hierarchical vocab-Agnostic Expert Clustering backbone to open-vocabulary panoptic segmentation, enabling highly computationally efficient training and inference. Finally, we demonstrated the first employment of an open-vocabulary panoptic segmentation model onto city-scale point clouds. Though this work promoted many promising advancements in the field of open scene understanding and foundation model based pseudo-labeling techniques, it also pinpointed several pressing areas for future inquiry in order to enable the production of a foundation scale model for the processing of large point clouds. We are optimistic that this research will push forward the state of 3D scene understanding. 

{
    \small
    \bibliographystyle{ieeenat_fullname}
    \bibliography{main}
}


\end{document}